# Do Benchmarks Underestimate LLM Performance? Evaluating Hallucination Detection With LLM-First Human-Adjudicated Assessment

Ismail Furkan Atasoy[1,*], Begum Mutlu[2], *Ebru Akcapinar Sezer*[1] and Abdelrahman Wahdan[3]

[1] *Department of Computer Engineering, Hacettepe University, Ankara, 06800, Türkiye*
[2] *Department of Computer Engineering, Ankara University, Ankara, 06830, Türkiye*
[3] *Zephlen AI and Information Technologies Inc., Kocaeli, 41400, Türkiye*

**Abstract**

Hallucination, also referred to as factual inconsistency, remains a persistent challenge in Large Language Models (LLMs), particularly in settings where models are expected to ground their outputs in provided context, such as RAG (Retrieval-Augmented Generation) and agentic AI systems. This study focuses on contextual hallucination detection in summarization tasks, which structurally resemble these systems. We analyze standardized versions of the QAGS-C and SummEval datasets by conducting a unified human adjudication, forming a disagreement-aware, adjudication-based evaluation framework that compares original benchmark annotations with LLM-generated reason and span-based hallucination detection predictions from Gemini 2.5 Flash and GPT-5 Mini. Our analysis reveals systematic divergences between human labels and LLM judgments. To address this, we re-evaluated all conflicted samples through a human adjudication process involving 2 cross-cultural adjudicators. The re-evaluation results in consistent improvements across both datasets, as reflected in notable increases in agreement and accuracy metrics. In particular, triple agreement (between the human, GPT, and Gemini) rates increased by 6.38% for QAGS-C and by approximately 7.62% for SummEval. Similarly, model accuracy improved after adjudication, with GPT accuracy increasing by 4.25% on QAGS-C and 2.34% on SummEval, while Gemini showed larger gains of 8.51% and 3.80%, respectively. Notably, when LLMs provided explicit reasoning and span-based evidence, adjudicators frequently sided with the model's judgments over the original human annotations. Overall, human adjudicator agreement ranged between 83% and 87%. These findings suggest that for ambiguity-prone tasks such as hallucination detection, single-pass annotations may be insufficient, and that iterative or model-assisted re-evaluation can yield more reliable and semantically grounded benchmarks.

**Keywords**

Large Language Models (LLM), Hallucination Detection, Factual Consistency, LLM-Assisted Adjudication, Benchmark Reliability, RAG Hallucination, Agent Hallucination

## 1. Introduction

Large language models (LLMs) generate text via probabilistic next-token prediction which may result in fluent and highly confident outputs that are factually incorrect or unsupported. In the literature, this case is widely referred to as hallucination. However, because hallucination is defined differently depending on the task (e.g., summarization, QA, data-to-text) and error type (e.g., unsupported claims, fabricated information), agreement on a single standardized hallucination definition remains limited. In practice, one of the accepted definitions of hallucination is generating information that is inconsistent with reality or not supported by the provided evidence [1, 2].

Hallucinations can be categorized based on the source of information they rely on [1]. Parametric hallucinations originate from a model's internalized knowledge, whereas contextual hallucinations arise when a model fails to correctly ground its output in the provided external source. Reliability concerns are increasingly prominent in retrieval-augmented and agentic generation systems, where hallucinations may arise even when the retrieved evidence itself is correct, due to errors in evidence selection, tool usage, or final response synthesis [3]. In this work, we focus on contextual hallucinations in abstractive summarization, where the source document is fully available and hallucination manifests as unsupported or contradictory content in the generated summary.

*Corresponding author.
ismailfurkanatasoy@hacettepe.edu.tr (I.F. Atasoy); bmbilge@ankara.edu.tr (B. Mutlu); ebru@hacettepe.edu.tr (E.A. Sezer); abdurrahman@zephlen.ai (A. Wahdan)
0009-0006-4183-5364 (I.F. Atasoy); 0000-0003-1960-2143 (B. Mutlu); 0000-0002-9287-2679 (E.A. Sezer); 0009-0004-0182-6514 (A. Wahdan)

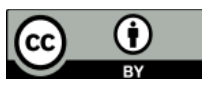

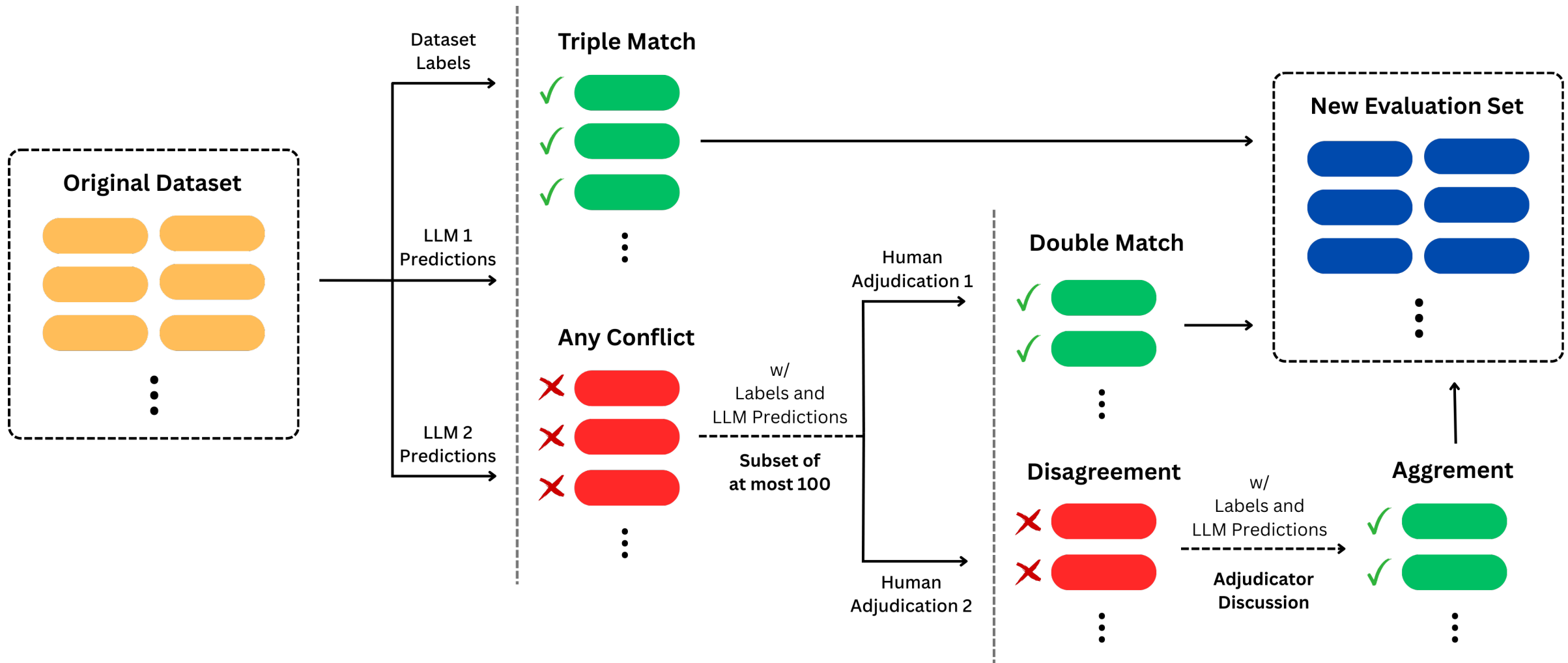

**Figure 1:** A diagram illustrating all stages of the re-evaluation process.

A common assumption behind the literature is that benchmark ground-truth labels (typically obtained through expert annotation or aggregated from multiple crowd annotators) serve as the final reference for correctness. However, this assumption is increasingly under-discussed, as recent work highlights substantial human label variation and cases where disagreement is inherent rather than noise [4]. Recent evidence suggests that frontier LLMs can achieve strong performance on several standardized academic and professional benchmarks, in some cases approaching human-level results [5]. This motivates a growing question: *Is evaluating state-of-the-art LLMs' hallucination-detection ability solely with existing hallucination benchmarks still adequate?*

To address this question, we leverage state-of-the-art LLMs to produce alternative hallucination judgments and span-level localizations of unsupported content, inspired by fine-grained annotation scheme [6]. We then conduct a unified human adjudication, forming a disagreement-aware, adjudication-based evaluation in which both the original benchmark annotations and the LLM-generated span-based labels. Each article-summary pair is processed by two LLMs, which independently generate span-based predictions. When the LLM outputs and the dataset label disagree, human adjudicators review the case to assign a final label. The resulting adjudicated examples constitute a new evaluation set (Fig. 1). This setup enables a direct comparison of model predictions and benchmark ground truth, revealing how often human annotators ultimately endorse LLM decisions at both the label and localization levels. Our findings collectively suggest that fixed benchmark-based evaluations are limited for hallucination detection studies. We advocate adopting an adjudication-based protocol, which allows these studies to be conducted and evaluated consistently across multiple datasets.

## 2. Related Work

Hallucination detection and factual consistency evaluation have been studied extensively in the context of abstractive summarization, which provides a clear setting where models must generate novel text while remaining faithful to the source. Summarization benchmarks offer both well-defined reference documents and the opportunity to evaluate errors at different granularities, making them a natural starting point for research in hallucination evaluation. Early work in this area focused on cases where hallucinations appear as content unsupported by the source or directly contradicting it. A variety of metrics have been proposed to capture such inconsistencies, including classification or natural language inference-based methods and question-answering-based approaches. For instance, FactCC [7] models text-level support between summary and source, while QA-based metrics such as QAGS and FEQA [8, 9] generate questions from the summary and verify answer consistency. SummaC [10] revisited NLI-based inconsistency detection specifically for summarization, highlighting the importance of mitigating granularity mismatches with NLI training data by reducing inputs to the sentence level. To address differences across summarization datasets and evaluation practices, TRUE [11] introduced a standardized meta-evaluation framework, enabling consistent interpretation of evaluation signals and their role in guiding targeted correction mechanisms.

While early benchmarks often treated hallucination as a binary phenomenon, recent work emphasizes the value of fine-grained annotations. Datasets such as RAGTruth [6] provide word-level human annotations distinguishing supported and unsupported spans, which allows for richer analyses and more interpretable evaluation. Building on this supervision, methods like LettuceDetect [12] formulate hallucination detection as a token-level classification task, explicitly localizing hallucinations in model outputs. Beyond detection, mitigation pipelines also benefit from fine-grained outputs. For example, RAG-HAT [13] leverages span-level hallucination predictions to guide a subsequent correction stage, typically executed by a stronger LLM, resulting in reduced unsupported claims. These approaches collectively motivate evaluation protocols that move beyond binary correctness to consider localization quality and downstream utility of span-level outputs.

More recently, the notion of "LLM-as-a-judge" has emerged, using large language models to provide scalable evaluation and supervision. TrueTeacher [14] employed an LLM to generate large-scale synthetic supervision for factual consistency evaluation, subsequently training smaller student evaluators on this supervision. In parallel, research on annotation reliability demonstrates that disagreement is systematic and can significantly affect datasets. Plank [4] argues that ground truth labels in NLP often reflect natural variation among human annotators, which should be modeled rather than suppressed. Empirically, Nahum et al. [15] showed that LLMs can detect likely label errors in benchmarks and that human re-annotation can substantially alter evaluation outcomes. This line of work underscores the risk of treating benchmark annotations as fixed oracles and motivates the integration of adjudication-based evaluation protocols.

The reason we chose summarization datasets for hallucination evaluations is that they allow for extracting information directly from the given context without requiring interpretation, making hallucinations easier to detect. Summarization benchmarks such as CNN/DailyMail, FactCC, QAGS, FEQA, and SummaC provide widely used metrics for measuring hallucination in generative models. Furthermore, RAG systems, which have become increasingly prominent in modern generative pipelines, often produce responses that summarize the provided context, making summarization datasets useful for RAG evaluation as well. For example, the RAGTruth benchmark includes a summarization subset alongside QA and data-to-text tasks, enabling the analysis of hallucination behavior in RAG outputs. Similarly, methods like LettuceDetect address RAG hallucinations while also leveraging summarization datasets.

## 3. Datasets

We conduct our experiments on SummEval [16] and QAGS-C [8], two benchmark datasets included in the TRUE Benchmark. The TRUE framework standardizes factual consistency evaluation across multiple summarization datasets by unifying annotation formats. Both SummEval and QAGS-C focus on document-grounded summarization, making them well-suited for studying contextual hallucinations, where unsupported or contradicted content arises despite full access to the source text.

**SummEval** is an evaluation dataset developed to assess multiple dimensions of summarization quality, including consistency, coherence, fluency, and relevance. It consists of news articles paired with summaries generated by a diverse set of summarization systems. Each summary is annotated by three expert and five crowd annotators, with each annotator assigning a score on a 1-5 Likert scale for each evaluation dimension. Within the TRUE framework, these graded consistency scores are standardized into binary factual consistency labels. Specifically, a summary is considered consistent (non-hallucinated) if and only if all three expert annotators assign a score of 5 for consistency. If any expert annotator assigns a score below 5, the summary is labeled as inconsistent (hallucinated).

**QAGS-C** is a factual consistency dataset derived from the QAGS framework and adapted and standardized as part of the TRUE framework. Similar to SummEval, it is based on pairs of original news documents and generated summaries, but it focuses exclusively on factual consistency. To facilitate annotation, each summary in QAGS-C is first segmented into individual sentences. Each sentence is then independently labeled by three expert annotators as either consistent “yes” or inconsistent “no” with respect to the source document. Unlike SummEval, TRUE applies a majority voting scheme to aggregate these annotations: a sentence is considered consistent if at least two out of three experts label it as “yes”. A summary is then labeled as consistent only if all of its sentences are consistent; otherwise, it is labeled as inconsistent.

In our experiments, we follow the exact same standardization protocol as TRUE and use the resulting binary labels without modification. Similar to RAG settings, where responses are generated conditioned on retrieved source documents, both datasets involve summaries generated directly from an available original document. While SummEval provides binary summary-level labels, QAGS-C offers sentence-level annotations, enabling a slightly more fine-grained view of factual consistency. This contrast allows us to study hallucination detection under different annotation granularities.

**Table 1**

Statistics for the datasets. The “Hallucination Rate” represents the ratio of inconsistent samples to the total number of samples in the dataset. “Article Length” refers to the total number of words in the original text from which the summary is derived.

| Dataset | No. of Samples | Hallucination Rate | Avg. Article Length | Avg. Summary Length |
|---|---|---|---|---|
| *SummEval* | 1600 | 18.4% | 359 | 63 |
| *QAGS-C* | 235 | 51.9% | 383 | 49 |

# 4. Method

## 4.1. LLM Inference

We selected two state-of-the-art LLMs, GPT-5 Mini[1] and Gemini 2.5[2] Flash, as evaluators. We obtained predictions from both models via a batch API call, retrieving structured outputs in JSON format. Given the ambiguity surrounding hallucination definitions, we designed a concise prompt instructing each model to perform hallucination and factual consistency checking with respect to a source document. For model parameters, we disabled “thinking outputs” or set them to the minimum possible level, and we left temperature at its default setting. Each model was expected to output a brief reason justifying that prediction, and a span indicating where any hallucinated content occurs when applicable or “none” if no hallucination is present. The reason field is included

[1] https://openai.com/tr-TR/index/introducing-gpt-5
[2] https://docs.cloud.google.com/vertex-ai/generative-ai/docs/models/gemini/2-5-flash

```
You are an expert in hallucination and factual inconsistency detection.

Task: Check if the following summary contains any hallucinations. Hallucination means information that is not supported by the article. Respond only in valid JSON, no explanations outside JSON.

### Article
{article}

### Summary
{summary}

### Output
Return a JSON object like:

```JSON
{
"reason" : "Detailed explanation of your judgment about hallucination/inconsistency",
"span" : "Exact hallucinated span from the summary (only the first one you can find, exact string match, as short as possible) or 'none' if there is no hallucination"
}
```
```

**Figure 2:** The prompt format used to obtain predictions from the LLMs.

to encourage the models to explicitly articulate their rationale and to guide their final decisions. The prompt structure is shown in Figure 2. Based on these outputs, we subsequently convert the responses into binary labels based on the values in the span field. If the span field contains any string other than “none”, the instance is considered hallucinated; otherwise, instances with “none” are treated as non-hallucinated.

## 4.2. Human Adjudication

As shown in Figure 1, samples for which there was no three-way agreement among the dataset label, GPT-5 Mini and Gemini 2.5 Flash were collected into a disagreement subset, capped at a maximum of 100 instances. These samples were independently re-evaluated by two human adjudicators who were blinded to each other’s decisions. The first and last authors who are from different countries, assumed to have different cultural backgrounds and natural language understanding capabilities actively participated this adjudication process. Both are from the field of computer science; one is an artificial intelligence engineer, and the other is a research assistant. Subsequently, the adjudication results were reviewed and approved by other authors, a professor and an assistant professor. They had no prior or concurrent experience related to either the specific dataset domains or the hallucination domain. This is partly because hallucination, by its nature, is not confined to any particular domain.

During the re-evaluation phase, each adjudicator was provided with the original dataset label as well as the reason and span outputs produced by both LLMs. Based on all available information, the adjudicators reviewed all conflict cases and had no additional information beyond the prompt given to the LLMs. They examined the examples based on their own definitions of hallucination and created new labels by filling in only the span field, relying on their zero-shot reasoning capabilities in a manner analogous to the LLMs.

After the independent annotation stage, all instances for which the two adjudicators disagreed were gathered for a second-round resolution. In this phase, the two adjudicators jointly reviewed the original dataset labels, the LLM predictions, and each other’s annotations and they discussed their views simply as two ordinary individuals, without undergoing any special preparation. If a consensus could not be reached, the final label was determined via majority voting over five labels: the original dataset label, the predictions from the two LLMs, and the two adjudicator labels. Otherwise, when one adjudicator adopted the other’s view, consensus was considered to be reached, and that view was accepted as the final label. At the end, the resulting labels constitute the final reference adjudications used in our evaluation.

**Table 2**

This table presents the agreement ratio and accuracy values based on the initial datasets. "Triple Aggrement" refers to aggrement cases where both LLMs and the dataset label share the same class label. "Dual Aggrement" refers to cases where the two LLMs agree with each other. "GPT Accuracy" and "Gemini Accuracy" indicate their respective accuracy performance with respect to the dataset labels. The values shown in bold indicate the LLM that achieves the best score for each dataset.

| Evaluation Metric | QAGSC | SummEval |
|---|---|---|
| *Triple Aggrement* | 82.55% | 84.75% |
| *Dual Aggrement* | 91.49% | 93.95% |
| *GPT Accuracy* | 85.96% | **88.66%** |
| *Gemini Accuracy* | **87.66%** | 86.89% |

# 5. Results

Once the responses obtained from the LLMs were converted into binary form, we computed several agreement rates with respect to the original dataset labels, as shown in Table 2. According to the initial findings, the dual agreement rate of the LLMs is approximately 9% higher than the three-way agreement rate. This indicates that there is roughly a 9% subset of examples for which both LLMs reach the same decision, but the human-provided dataset label does not align with them.

After adjudication on a maximum of 100 instances (all 41 such cases from QAGS-C and the first 100 out of 242 cases from SummEval), the agreement between the two human adjudicators reached 87.80% for QAGS-C and 83% for SummEval. Using these adjudicated labels as reference, the examples selected as hallucinated or consistent by the LLMs were subsequently evaluated separately for each dataset, with the results reported in Table 3. GPT predicted relatively few hallucinations, whereas Gemini produced substantially more hallucination predictions. In conflict cases between dataset labels and model predictions, the adjudicators preferred hallucination judgments at rates of 82.93% for QAGS-C and 73.00% for SummEval, indicating a general tendency to favor hallucination labels when there is any indication of inconsistency. Despite Gemini's higher propensity to flag hallucinations, the human acceptance rate for GPT in such "alone" conflict cases was 8.33% for QAGS-C and 30.00% for SummEval. For Gemini, the corresponding endorsement rates were substantially higher, at 74.98% for QAGS-C and 74.07% for SummEval. When both models agreed with each other but disagreed with the original dataset annotations, adjudicators sided with the models at rates of 71.41% for QAGS-C and 87.30% for SummEval.

**Table 3**

This table presents the results after the adjudicator evaluation. Under the "Group" heading, "GPT" and "Gemini" represents the samples where corresponding LLM alone caused a conflict. The group named "Both" reflects cases where both LLMs agreed with each other, but the dataset label disagreed with them. The "Total" value represents the proportion of the corresponding row and column's total examples relative to the total number of conflict samples, while "Adj. Selection" represents the proportion of examples selected by the adjudicators relative to the total number of examples.

| | QAGSC (41 Conflict Samples) | | | | SummEval (100 Coflict Samples) | | | |
|---|---|---|---|---|---|---|---|---|
| | Hallucinated | | Consistent | | Hallucinated | | Consistent | |
| Group | Adj. Selection | Total | Adj. Selection | Total | Adj. Selection | Total | Adj. Selection | Total |
| *GPT* | 2.44% | 2.44% | 0.00% | 26.83% | 2.00% | 2.00% | 1.00% | 8.00% |
| *Gemini* | 14.63% | 14.63% | 0.00% | 4.88% | 20.00% | 25.00% | 0.00% | 2.00% |
| *Both* | 21.95% | 24.39% | 14.63% | 26.83% | 34.00% | 34.00% | 21.00% | 29.00% |

In Table 4, the newly adjudicated labels are merged with the original datasets, resulting in an updated version of Table 2. In this merging process, the initial annotations of triple-agreement samples are treated as ground truth, and only the adjudicated labels of conflict samples are replaced. While all conflict samples in QAGS-C were included in the adjudication process, only the first 100 conflict samples in SummEval were adjudicated. For the remaining 142 samples, it is assumed that they follow the statistics of the evaluated subset of 100 samples. The results show that LLM accuracy consistently increases under the new labels. Moreover, the triple-agreement scores approach the dual-agreement scores, substantially narrowing the gap between LLMs and human labels.

**Table 4**

This table presents the updated version of Table 2, constructed by replacing the original annotations in the dataset with the new adjudications. The "~" symbol indicates estimated statistics for samples that did not undergo the adjudication process, extrapolated based on the proportion of adjudicated conflict samples. Values in parentheses represent the differences with respect to Table 2. The values shown in bold indicate the LLM that achieves the best score for each dataset.

| Evaluation Metric | QAGSC | SummEval |
|---|---|---|
| *Triple Aggrement* | 88,93% (+6,38%) | ~92.37% (+7,62%) |
| *Dual Aggrement* | 91.49% (-) | 93.95% (-) |
| *GPT Accuracy* | 90,21% (+4,25%) | **~91.00%** (+2,34%) |
| *Gemini Accuracy* | **96,17%** (+8,51%) | ~90.69% (+3,80%) |

## 6. Discussion and Conclusion

An important observation concerns the change in hallucination prevalence after adjudication. Although the original SummEval dataset contained only 18.4% hallucinated examples, the new combined dataset marked 20.5% of the examples as hallucinated. A similar trend was observed for QAGS-C, where the proportion of hallucinated examples increased from 51.9% to 56.17% after adjudication. When disagreements arose between human annotators and LLMs, especially in cases where LLMs provided detailed hallucination reasoning and span-based evidence, adjudicators tended to side with the LLMs in most cases. This suggests that LLMs can capture subtle factual inconsistencies that may be overlooked by human annotators, even in datasets constructed via majority voting from multiple expert annotations.

One possible factor influencing these results is annotation granularity. SummEval provides only binary summary level labels, whereas QAGS-C offers sentence level supervision. Our findings suggest that exposure to LLM-provided reasons and hallucination spans can meaningfully influence adjudicators' decisions, particularly in datasets with coarse-grained labels. This raises the possibility that richer, span and reason-based annotations could reduce ambiguity, improve agreement, and lead to more faithful hallucination assessments. Future work could validate this hypothesis using datasets such as RAGTruth, which include span level annotations accompanied by explicit reasoning.

While our experiments focus on summarization benchmarks, the underlying phenomenon is unlikely to be task specific. Hallucination is also a critical concern in settings such as question answering, retrieval augmented generation, and agentic AI systems, where errors may propagate across multiple reasoning steps or actions. Extending disagreement aware adjudication protocols to these domains may reveal similar underestimation effects and further demonstrate the value of LLM assisted evaluation beyond summarization. This is particularly important given the growing interest in using LLMs as judges for hallucination detection, a direction that we expect to see rapid adoption in related studies. Addressing this problem more extensively will require broader engagement from the research community, as well as the development of additional fine-grained datasets that include both LLM and human-generated annotations, such as reason and span-based labels.

Another key observation concerns the lack of a universally accepted definition of hallucination. The high number of conflicts observed between LLMs and human annotators, as well as among adjudicators themselves, appears partly attributable to differing internal interpretations of what constitutes hallucination. GPT's tendency to produce more consistent but conservative outputs may reflect a narrower or more rigid conceptualization of hallucination, whereas Gemini appears to adopt a broader interpretation that aligns more closely with human judgments when supported by explanations. These findings suggest that more detailed prompts and a clearly articulated hallucination framework could improve alignment and consistency across both human and model judgments. In addition, we did not identify any specific patterns or systematic observations regarding cases where the LLMs diverged from the initial dataset annotations, or the conditions under which the adjudicators favored the LLM outputs. However, we think that this aspect warrants further investigation in future work.

Nevertheless, based on our results, we suggest that for tasks such as hallucination detection, which rely heavily on nuanced natural language understanding and lack a sharply defined ground truth, treating a single annotation pass as definitive can be misleading. Incorporating disagreement analysis, adjudication, and LLM generated explanations offers a more robust alternative, enabling both improved evaluation fidelity and deeper insight into the nature of hallucinated content.

## Declaration on Generative AI

During the preparation of this work, the authors used ChatGPT and Gemini in order to: Grammar and spelling check. After using these services, the authors reviewed and edited the content as needed and take full responsibility for the publication's content.